\documentclass{llncs}

\usepackage[english]{babel}
\usepackage[utf8]{inputenc}
\usepackage{mathrsfs,amsmath}
\usepackage{amsfonts}
\usepackage{graphicx}
\usepackage[colorinlistoftodos]{todonotes}
\usepackage{hyperref}
\usepackage{multirow}
\usepackage{amssymb}
\usepackage{hhline}
\usepackage[tmargin=2cm,bmargin=2cm,lmargin=2cm,rmargin=2cm]{geometry}
\usepackage{enumitem}
\usepackage{subcaption}
\usepackage{breqn}
\usepackage{placeins}
\usepackage{array}
\usepackage{booktabs}
\usepackage{listings}
\usepackage{float}
\usepackage[]{scrextend}
\usepackage[]{upgreek}
\usepackage{color,soulutf8}

\begin{document}
\title{Impact of Sentiment Detection to Recognize Toxic and Subversive Online Comments}

\author{Éloi Brassard-Gourdeau \and Richard Khoury}

\institute{Department of Computer Science and Software Engineering, Université Laval, Quebec City, Canada \email{eloi.brassard-gourdeau.1@ulaval.ca richard.khoury@ift.ulaval.ca}}

\maketitle

\begin{abstract}
The presence of toxic content has become a major problem for many online communities. Moderators try to limit this problem by implementing more and more refined comment filters, but toxic users are constantly finding new ways to circumvent them. Our hypothesis is that while modifying toxic content and keywords to fool filters can be easy, hiding sentiment is harder. In this paper, we explore various aspects of sentiment detection and their correlation to toxicity, and use our results to implement a toxicity detection tool. We then test how adding the sentiment information helps detect toxicity in three different real-world datasets, and incorporate subversion to these datasets to simulate a user trying to circumvent the system. Our results show sentiment information has a positive impact on toxicity detection against a subversive user.
\end{abstract}


\section{Introduction}
\label{sec:introduction}
Online communities abound today, forming on social networks, on webforums, within videogames, and even in the comments sections of articles and videos. While this increased international contact and exchange of ideas has been a net positive, it has also been matched with an increase in the spread of high-risk and toxic content, a category which includes cyberbullying, racism, sexual predation, and other negative behaviors that are not tolerated in society. The two main strategies used by online communities to moderate themselves and stop the spread of toxic comments are automated filtering and human surveillance. However, given the sheer number of messages sent online every day, human moderation simply cannot keep up, and either leads to a severe slowdown of the conversation (if messages are pre-moderated before posting) or allows toxic messages to be seen and shared thousands of times before they are deleted (if they are post-moderated after being posted and reported). In addition, human moderation cannot scale up easily to the number of messages to monitor; for example, Facebook has a team of 20,000 human moderators, which is both massive compared to the total of 25,000 other employees in the company, and minuscule compared to the fact its automated algorithms flagged messages that would require 180,000 human moderators to review\footnote{http://fortune.com/2018/03/22/human-moderators-facebook-youtube-twitter/}. Keyword detection, on the other hand, is instantaneous, scales up to the number of messages, and prevents toxic messages from being posted at all, but it can only stop messages that use one of a small set of denied words, and, are thus fairly easy to circumvent by introducing minor misspellings (i.e. writing "kl urself" instead of "kill yourself"). In \cite{subversion}, the authors show how  minor changes can elude even complex systems. These attempts to bypass the toxicity detection system are called subverting the system, and toxic users doing it are referred to as subversive users.

In this paper, we consider an alternative strategy for toxic message filtering. Our intuition is that, while toxic keywords can easily be disguised, the toxic emotional tone of the message cannot. Consequently, we will study the correlation between sentiment and toxicity and its usefulness for toxic message detection both in subversive and non-subversive contexts.

The rest of this paper is structured as follows. After a review of the relevant literature in the next section, we will consider the problem of sentiment detection in online messages in Section \ref{sec:sentiment}. Next, we will study the measure of toxicity and its correlation to message sentiment in Section \ref{sec:toxicity}. Finally, we will draw some concluding remarks in Section \ref{sec:conclusion}.

\section{Related Work}
\label{sec:background}

Given the limitations of human and keyword-based toxicity detection systems mentioned previously, several authors have studied alternative means of detecting toxicity. In one of the earliest works on the detection of hate speech, the authors of \cite{warner2012detecting} used n-grams enhanced by part-of-speech information as features to train an SVM classifier to accurately pick out anti-semitic online messages. Following a similar idea, the authors of \cite{nobata2016abusive} conducted a study of the usefulness of various linguistic features to train a machine learning algorithm to pick out hate speech. They found that the most useful single feature was character n-grams, followed closely by word n-grams. However, it was a combination of all their features (n-grams, features of language, features of syntax, and word embedding vectors) that achieved the highest performance. The authors of \cite{alorainy2018cyber} studied hate speech through the detection of othering language. They built a custom lexicon of pronouns and semantic relationships in order to capture the linguistic differences when describing the in-group and out-group in messages, and trained a word embedding model on that data. 

Hate speech is not the only form of toxicity that has been studied. In \cite{reynolds2011using}, the authors studied cyberbullying. They developed a list of 300 "bad" words sorted in five levels of severity. Next, they used the number and density of "bad" words found in each online message as the features to train a set of machine learning systems. The authors of \cite{ebrahimi2016automatic} also used words as featured in two systems, this time to detect sexual predators. One used the TFxIDF values of the words of the text to train a single-class SVM classifier, and the other used a bag-of-words vector of the text as input to a deep neural network. The authors found that the latter system offered the better performance in their experiments. 

Recently, deep learning has become very popular for NLP applications, and pre-trained word embeddings have been shown to be very effective in most text-based neural network applications. In \cite{deepcyber2018}, four different deep learning models were implemented and shown to outperform benchmark techniques for cyberbullying detection on three different datasets. In \cite{meanbirds}, a deep neural network taking a word embedding vector as input was used to detect cyberbullying on Twitter. 

It thus appears from the related literature that authors have tried a variety of alternative features to automatically detect toxic messages without relying strictly on keyword detection. However, sentiment has rarely been considered. It was one of the inputs of the deep neural network of \cite{meanbirds}, but the paper never discussed its importance or analyzed its impact. The authors of \cite{vanhee2018} conducted the first study of cyberbullying in Dutch, and considered several features, including a subjectivity keyword lexicon. They found its inclusion helped improve results, but that a more sophisticated source of information than simple keyword detection was required. And the study of \cite{sentimentcyber} used the sentiment of messages, as measured by the SentiStrength online system, as one of several features to detect cyberbullying messages. However, an in-dept analysis of how sentiment can benefit toxicity detection has not been done in any of these papers, and a study of the use of sentiment in a subversive context has never been done.

\section{Sentiment Detection}
\label{sec:sentiment}

\subsection{Lexicons}
Sentiment detection, or the task of determining whether a document has a positive or negative tone, has been frequently studied in the literature. It is usually done by using a sentiment lexicon that either classifies certain words as positive or negative, or quantifies their level of positivity or negativity. We decided to consider six such lexicons:
\begin{itemize}
  \item \textbf{SentiWordNet}\footnote{http://sentiwordnet.isti.cnr.it/} is a widely-used resource for sentiment mining. It is based on WordNet, and assigns three scores to each synset, namely positivity, negativity, and objectivity, with the constraint that the sum of all three must be 1. Using this lexicon requires a bit of preprocessing for us, since the same word can occur in multiple different synsets with different meanings and therefore different scores. Since picking out the intended meaning and synset of a polysemous word found in a message is beyond our scope, we instead chose to merge the different meanings and compute a weighted average of the scores of the word. The weights are the ranks of the synsets, which correspond to the popularity of that meaning of the word in documents. The average score equation is :
\begin{equation}
score = \frac{\sum^k \frac{score}{rank}}{\sum^k \frac{1}{rank}}
\end{equation}
where $k$ is the number of times the word occurs with the same part of speech. We compute the average positivity and negativity scores, but not the objectivity scores, since they are not useful for our purpose and since they are simply the complement of the other two. This allows us to extract 155,287 individual words from the lexicon, with a positivity and negativity score between 0 and 1 for each. We should note that SentiWordNet differentiates a word based on part-of-speech, and we maintain this distinction in our work

  \item \textbf{Afinn}\footnote{https://github.com/fnielsen/afinn} is a lexicon of 3,382 words that are rated between -5 (maximum negativity) and 5 (maximum positivity). To match SentiWordNet, we split this score into positivity and negativity scores between 0 and 1. For example, a word with a $-3$ score was changed to have a positive score of $0$ and a negative score of $0.6$. 
  
  \item \textbf{Bing Liu}\footnote{https://www.cs.uic.edu/~liub/FBS/sentiment-analysis.html} compiled lists of 6,789 positive or negative words. Given no other information, we assigned each word in the positive list a positivity score of 1 and a negativity score of 0, and vice-versa for the negative-list words. 
  
  \item \textbf{General Inquirer} \footnote{http://www.wjh.harvard.edu/~inquirer/} is a historically-popular lexicon of 14,480 words, though only 4,206 of them are tagged as either positive or negative. As for the Bing Liu lexicon, we assigned binary positive and negative scores to each word that was tagged as positive or negative.

  \item \textbf{Subjectivity Clues}\footnote{http://mpqa.cs.pitt.edu/lexicons/} extends the sentiment tags of the General Inquirer up to 8,222 words using a dictionary and thesaurus. It also adds a binary strength level (strong or weak) to the polarity information. We merged polarity and strength as a measure of 0.5 and 1 for weak or strong positivity or negativity.
  
  \item \textbf{NRC}\footnote{https://www.nrc-cnrc.gc.ca/eng/rd/ict/emotion_lexicons.html} has a list of 14,182 words that are marked as associated (1) or not associated (0) with 8 emotions (anger, fear, anticipation, trust, surprise, sadness, joy, disgust) and two sentiments (negative and positive). We transform this association into binary positive and negative scores in the same way we did for Bing Liu and General Inquirer.
  
\end{itemize}

All six of these lexicons have limitations, which stem from their limited vocabulary and the ambiguity of the problem. Indeed, despite being thousands of words each and covering the same subject and purpose, our six lexicons have only 394 words in common, indicating that each is individually very incomplete compared to the others. And we can easily find inconsistencies between the ratings of words, both internally within each lexicon and externally when we compare the same words between lexicons. Table \ref{tbl:lexicons} illustrate some of these inconsistencies: for instance, the word "helpless" is very negative in SentiWordNet but less so in Afinn and Subjectivity Clues, while the word "terrorize" is more strongly negative in the latter two resources but less negative (and even a bit positive) in SentiWordNet. Likewise, the word "joke" is strongly positive, weakly positive, or even negative, depending on the lexicon used, and the word "merry" is more positive than "joke" according to every lexicon except SentiWordnet, which rates it equally positive and negative. By contrast the word "splendid" has the same positivity values as "merry" in all lexicons except SentiWordnet, where it has the highest possible positivity score. In a longer document, such as the customer reviews these lexicons are typically used on \cite{ohana2012case,tumsare2014opinion,agarwal2015sentiment}, these problems are minor: the abundance and variety of vocabulary in the text will insure that the correct sentiment emerges overall despite the noise these issues cause. This is not true for the short messages of online conversations, and it has forced some authors who study the sentiments of microblogs to resort to creating or customizing their own lexicons \cite{nielsen2011new}. This, incidentally, is also why we could not simply use an existing sentiment classifier. We will instead opt to combine these lexicons into a more useful resource.

\begin{table}	
\caption{Sentiment of words per lexicon}
\label{tbl:lexicons}
\begin{tabular}{l|c|r|c|c|c|c}
\hline
Word & SentiWordNet & Afinn & Bing Liu & General Inquirer & Subjectivity Clues & NRC \\ \hhline{=|=|=|=|=|=|=}
terrorize & ['0.125', '0.250'] & -3 & negative & negative & strong negative & negative \\
helpless & ['0.000', '0.750'] & -2 & negative & negative & weak negative & negative \\
joke & ['0.375', '0.000'] & 2 & negative & positive & strong positive & negative \\
merry & ['0.250', '0.250'] & 3 & positive & positive & strong positive & positive \\
splendid & ['1.000', '0.000'] & 3 & positive & positive & strong positive & positive \\
\hline
\end{tabular}
\end{table}

\subsection{Message Preprocessing}
The first preprocessing step is to detect the presence and scope of negations in a message. Negations have an important impact; the word "good" may be labeled positive in all our lexicons, but its actual meaning will differ in the sentences "this movie is good" and "this movie is not good". We thus created a list of negation keywords by combining together the lists of the negex algorithm\footnote{https://github.com/mongoose54/negex/tree/master/ negex.python} and of \cite{rulebasednegationscope}, filtering out some irrelevant words from these lists, and adding some that were missing from the lists but are found online.

Next, we need to determine the scope of the negation, which means figuring out how many words in the message are affected by it. This is the challenge of, for example, realizing that the negation affects the word "interesting" in "this movie is not good or interesting" but not in "this movie is not good but interesting". We considered two algorithms to detect the scope of negations. The first is to simply assume the negation affects a fixed window of five words\footnote{The average window size in our test dataset was 5.36 words, so we rounded to the closest integer.} after the keyword \cite{councill2010s}, while the second discovers the syntactic dependencies in the sentence in order to determine precisely which words are affected \cite{dadvarscope}.

We tested both algorithms on the SFU review corpus of negation and speculation\footnote{https://www.researchgate.net/publication/256766329\_ SFU\_Review\_Corpus\_Negation\_Speculation}. As can be seen in Table \ref{tbl:negationdetection} the dependency algorithm gave generally better results, and managed to find the exact scope of the negation in over 43\% of sentences. However, that algorithm also has a larger standard deviation in its scope, meaning that when it fails to find the correct scope, it can be off by quite a lot, while the fixed window is naturally bounded in its errors. Moreover, the increased precision of the dependencies algorithm comes at a high processing cost, requiring almost 30 times longer to analyze a message as the fixed window algorithm. Given that online communities frequently deal with thousands of new messages every second, efficiency is a major consideration, and we opted for the simple fixed window algorithm for that reason.

\begin{table}
\caption{Comparison between fixed window and syntactic dependencies negation detection algorithms}
\begin{tabular}{l|c|c}
\hline
 & Fixed window & Dependencies\\ 
 \hhline{=|=|=}
Accuracy & 71.75\% & 82.88\% \\
Recall & 95.48\% &  90.00\%\\
Precision & 69.65\% & 78.37\% \\
Exact match & 9.03\% & 43.34\% \\
Standard deviation & 3.90 words & 5.54 words\\
ms per sentence & 2.4 & 68\\
\hline
\end{tabular}
\label{tbl:negationdetection}
\end{table}

The second preprocessing step is to detect sentiment-carrying idioms in the messages. For example, while the words "give" and "up" can both be neutral or positive, the idiom "give up" has a clear negative sentiment. Several of these idioms can be found in our lexicons, especially SentiWordNet (slightly over $60,000$). We detect them in our messages and mark them so that our algorithm will handle them as single words going forward.

Finally, we use the NLTK wordpunkt\_tokenizer to split sentences into words, and the Stanford fasterEnglishPOSTagger to get the part-of-speech of each word. Since our lexicons contain only four parts-of-speech (noun, verb, adverb, and adjective) and Stanford's tagger has more than 30 possible tags, we manually mapped each tag to one of the four parts-of-speech (for example, "verb, past participle" maps to "verb").

\subsection{Message Sentiment}
Once every word has a positivity and a negativity score, we can use them to determine the sentiment of an entire message. We do this by computing separately the sum of positive scores and of negative scores of words in the message, and subtracting the negative total from the positive total. In this way, a score over 0 means a positive message, and a score under 0 means a negative message. We consider two alternatives at this point: one in which we sum the sentiment value of all words in the sentence, and one where we only sum the sentiment value of the  top-three words with the highest scores for each polarity. We label these "All words" and "Top words" in our results. The impact of this difference is felt when we consider a message with a few words with a strong polarity and a lot of words with a weak opposite polarity; in the "Top words" scheme these weak words will be ignored and the strong polarity words will dictate the polarity of the message, while in the "All words" scheme the many weak words can sum together to outweigh the few strong words and change the polarity of the message.

We optionally take negations into account in our sentiment computation. When a word occurs in the word window of a negation, we flip its positivity and negativity scores. In other words, instead of adding its positivity score to the positivity total of the sentence, we added its negativity score, and the other way round for the negativity total. Experiments where we do that are labeled "Negativity" in our results. 

Finally, we optionally incorporate word weights based on their frequency in our datasets. When applied, the score of each word is multiplied by a frequency modifier, which we adapted from \cite{ohana2012case}:
\begin{equation}
frequency\_modifier = 1 - \sqrt[]{\frac{n}{n_{max}}}
\end{equation}
where $n$ is the number of times the word appears in a dataset, and $n_{max}$ is the number of times the most frequent word appears in that dataset. Experiments using this frequency modifier are labeled "Frequency" in our results.

\subsection{Experimental Results}
Our experiments have four main objectives: (1) to determine whether the "All words" or the "Top words" strategy is preferable; (2) to determine whether the inclusion of "Negation" and "Frequency" modifiers is useful; (3) to determine which of the six lexicons is most accurate; and (4) to determine whether a weighted combination of the six lexicons can outperform any one lexicon. 

To conduct our experiments, we used the corpus of annotated news comments available from the Yahoo Webscope program\footnote{Dataset L32: https://webscope.sandbox.yahoo.com/catalog.php ?datatype=l}. The comments in this dataset are annotated by up to three professional, trained editors to label various attributes, including type, sentiment and tone. Using these three attributes, we split the dataset into two categories, sarcastic and non-sarcastic, and then again into five categories, clear negative, slight negative, neutral, slight positive, and clear positive. Finally, we kept only the non-sarcastic comments where all annotators agreed to reduce noise. This gives us a test corpus of 2,465 comments.

To evaluate our results, we compute the sentiment score of each comment in our test corpus using our various methods, and we then compute the average sentiment score of comments in each of the five sentiment categories. For ease of presentation, we give a simplified set of results in Table \ref{tbl:sentimentresults1}, with only the average score of the two negative and the two positive labels combined, along with the overlap of the two distributions. The overlap is obtained by taking two normal distributions with the the means and standard deviations of the positive and the negative sets, and calculating the area in common under both curves. It gives us a measure of the ambiguous region where comments may be positive or negative. A good sentiment classifier will thus have very distant positive and negative scores and a very low overlap.

These results show that there are important differences between the lexicons. Three of the six are rather poor at picking out negative sentiments, namely Subjectivity Clues (where negative sentences are on average detected as more positive than the positive sentences), General Inquirer, and NRC. This bias for positivity is an issue for a study on toxicity, which we expect to be expressed using negative sentiments. The other three lexicons give a good difference between positive and negative sentences. For these three lexicons, we find that using \textit{All words} increases the gap between positive and negative sentence scores but greatly increases the standard deviation of each sentiment class, meaning the sentiment of the messages becomes ambiguous. On the other hand, using \textit{Top words} reduces the overlap between the distributions and thus gives a better separation of positive and negative sentiments. And while adding frequency information or negations does not cause a major change in the results, it does give a small reduction in overlap. 

\begin{table*}
\caption{Average sentiment scores of negative and positive (respectively) labeled sentences, and their overlap.}
\begin{tabular}{l|c|c|c|c|c|c}
\hline
Experiment & SWN & Afinn & Bing Liu & Gen. Inquirer & Subj. Clues & NRC \\
 \hhline{=|=|=|=|=|=|=}
All words & [-0.22, 0.31] 0.81 & [-0.43, 0.45] 0.71 & [-1.17, 0.69] 0.67 & [ 0.03, 1.44] 0.73 & [2.31, 1.97] 0.76 & [-0.15, 1.00] 0.77 \\
All + Negation & [-0.34, 0.17] 0.79 & [-0.44, 0.39] 0.69 & [-1.08, 0.61] 0.70 & [-0.27, 0.99] 0.77 & [1.66, 1.52] 0.83 & [-0.62, 0.75] 0.75\\
All + Frequency & [-0.21, 0.29] 0.80 & [-0.42, 0.40] 0.71 & [-1.17, 0.58] 0.68 & [-0.09, 1.23] 0.76 & [1.98, 1.70] 0.82 & [-0.19, 0.90] 0.79\\
All + Neg. + Freq. & [-0.29, 0.18] 0.78 & [-0.42, 0.35] 0.69 & [-1.06, 0.52] 0.71 & [-0.33, 0.85] 0.79 & [1.45, 1.34] 0.86 & [-0.56, 0.69] 0.77\\
Top words & [-0.23, 0.11] 0.75 & [-0.23, 0.31] 0.68 & [-0.54, 0.54] 0.67 & [-0.03, 0.59] 0.80 & [1.18, 1.17] 0.99 & [-0.14, 0.54] 0.77\\
Top + Negation & [-0.24, 0.10] 0.74 & [-0.24, 0.29] 0.67 & [-0.50, 0.53] 0.67 & [-0.12, 0.57] 0.77 & [0.86, 0.71] 0.94 & [-0.28, 0.49] 0.73\\
Top + Frequency & [-0.16, 0.15] 0.74 & [-0.23, 0.28] 0.67 & [-0.56, 0.47] 0.67 & [-0.07, 0.52] 0.79 & [1.00, 1.01] 0.99 & [-0.15, 0.50] 0.77\\
Top + Neg. + Freq. & [-0.17, 0.14] 0.73 & [-0.23, 0.26] 0.67 & [-0.51, 0.48] 0.66 & [-0.14, 0.49] 0.77 & [0.61, 0.76] 0.93 & [-0.26, 0.45] 0.74\\
\hline
\end{tabular}
\label{tbl:sentimentresults1}
\end{table*}

To study combinations of lexicons, we decided to limit our scope to SentiWordNet, Afinn, and Bing Liu, the three lexicons that could accurately pick out negative sentiments, and on the \textit{Top words} strategy. We consider three common strategies to combine the results of independent classifiers: majority voting, picking the one classifier with the maximum score (which is assumed to be the one with the highest confidence in its classification), and taking the average of the scores of all three classifiers. For the average, we tried using a weighted average of the lexicons and performed a grid search to find the optimal combination. However, the best results were obtained when the three lexicons were taken equally. For the majority vote, we likewise take the average score of the two or three classifiers in the majority sentiment.

Table \ref{tbl:sentimentresults2} presents the results we obtained with all three strategies. It can be seen that combining the three classifiers outperforms taking any one classifier alone, in the sense that it creates a wider gap between the positive and negative sentences and a smaller overlap. It can also be seen that the addition of negation and frequency information gives a very small improvement in the results in all three cases. Comparing the three strategies it can be seen that the maximum strategy is the one with the biggest gap in between positive and negative distribution, which was to be expected since the highest positive or negative sentiment is selected each time while it gets averaged out in the other two classifiers. However, the average score strategy creates a significantly smaller standard deviation of sentiment scores and a lower overlap between the distributions of positive and negative sentences. For that reason, we find the average score to be the best of the three combination strategies.

\begin{table*}
\caption{Sentiment scores using combinations of lexicons.}
\begin{tabular}{l|c|c|c}
\hline
Experiment & Majority vote & Maximum wins & Average scores \\
 \hhline{=|=|=|=}
Top words & [-0.36, 0.34] 0.67 & [-0.60, 0.52] 0.67  & [-0.32, 0.32] 0.64 \\
Top + Negation & [-0.35, 0.34] 0.66 & [-0.59, 0.51] 0.66  & [-0.31, 0.30] 0.63 \\
Top + Frequency & [-0.34, 0.32] 0.66  & [-0.58, 0.48] 0.67  & [-0.31, 0.30] 0.63 \\
Top + Neg. + Freq. & [-0.32, 0.30] 0.65  & [-0.55, 0.50] 0.65  & [-0.29, 0.29] 0.63 \\
\hline
\end{tabular}
\label{tbl:sentimentresults2}
\end{table*}
 
In all cases, we find that most misclassified sentences in our system are due to the lack of insults in the vocabulary. For example, none of the lexicons include colorful insults like "nut job" and "fruitcake", so sentences where they appear cannot be recognized as negative. Likewise, some words, such as the word "gay", are often used as insults online, but have positive meanings in formal English; this actually leads to labeling insult messages as positive sentences. This issue stems from the fact that these lexicons were designed for sentiment analysis in longer and more traditional documents, such as customer reviews and editorials. One will seldom, if ever, find insults (especially politically-incorrect ones such as the previous examples) in these documents.

\section{Toxicity Detection}
\label{sec:toxicity}

The main contribution of this paper is to study how sentiment can be used to detect toxicity in subversive online comments. To do this, we will use three new test corpora:
\begin{itemize}
  \item The \textbf{Reddit}\footnote{https://bigquery.cloud.google.com/table/fh-bigquery:reddit_comments.2007?pli=1} dataset is composed of over 880,000 comments taken from a wide range of subreddits and annotated a few years ago by the \textit{Community Sift} tool developed by \textit{Two Hat Security}\footnote{https://www.twohat.com/community-sift/}. This toxicity detection tool, which was used in previous research on toxicity as well \cite{mohan2017impact}, uses over 1 million n-gram rules in order to normalize then categorize each message into one of seven toxicity levels, detailed in Table \ref{tbl:commsift}, for a wide array of different categories. In our case, we consider the scores assigned to each message in five categories, namely bullying, fighting, sexting, vulgarity, and racism.
  \item The \textbf{Wikipedia Talk Labels}\footnote{https://figshare.com/articles/Wikipedia\_Talk\_Labels\_Toxicity /4563973} dataset consists of over 100,000 comments taken from discussions on English Wikipedia's talk pages. Each comment was manually annotated by around ten Crowdflower workers as toxic or not toxic. We use the ratio of toxic marks as a toxicity score. For example, if a sentence is marked toxic by 7 out of 10 workers, it will have a 0.7 toxicity score.
   \item The \textbf{Kaggle toxicity competition}\footnote{https://www.kaggle.com/c/jigsaw-toxic-comment-classification-challenge} dataset is also taken from discussions on English Wikipedia talk pages. There are approximatively 160,000 comments, which were manually annotated with six binary labels: toxic, severe\_toxic, obscene, threat, insult, and identity\_hate. This allows us to rate comments on a seven-level toxicity scale, from 0/6 labels marked to 6/6 labels marked. 
\end{itemize}

\begin{table*}
\caption{Toxicity levels in Community Sift.}
\begin{tabular}{l|l|l}
\hline
Level & Name & Description \\
 \hhline{=|=|=}
0 & Super-safe & Common safe words and n-grams, such as "hello".\\
1 & Whitelist & Words and n-grams manually marked as safe. \\
2 & Grey & Words and n-grams that could be combined with others to create unsafe messages.\\
3 & Questionable & Words and n-grams that may be unsafe, depending on context.\\
4 & Unknown & Spelling mistakes and unknown words.\\
5 & Mild & Controversial words and n-grams that may be safe, depending on context.\\
6 & Bad & Words and n-grams that are typically unsafe.\\
7 & Severe & High-risk n-grams (not words).\\ 
\hline
\end{tabular}
\label{tbl:commsift}
\end{table*}
\subsection{Correlation}
Our first experiment consists in computing the sentiment of each message in each of our three test corpora, and verifying how they correlate with the different toxicity scores of each of the corpora. Following the results we found in Section \ref{sec:sentiment}, we used the best three lexicons (SentiWordNet, Afinn, and Bing Liu), combined them by taking the average score, and used our four algorithm variations. The results are presented in Table \ref{tbl:correlations}. It can be seen that there is a clear negative correlation between toxicity and sentiment in the messages, as expected. Our results also show that using words only or including frequency information makes the relationship clearer, while adding negations muddies it. These results are consistent over all three test corpora, despite being from different sources and labeled using different techniques. The lower score on the Reddit dataset may simply be due to the fact it was labeled automatically by a system that flags potentially dangerous content and not by human editors, so its labels may be noisier. For example, mentioning sexual body parts will be labeled as toxicity level 5 even if they are used in a positive sentence, because they carry more potential risk.

\begin{table}
\caption{Correlation between sentiment and toxicity.}
\begin{tabular}{l|c|c|c}
\hline
Sentiment detection & Reddit & Wikipedia & Kaggle\\
 \hhline{=|=|=|=}
Top words & -0.2410 &  -0.3839 & -0.3188\\
Top + Negation & -0.2021 & -0.3488 & -0.2906\\
Top + Frequency & -0.2481 & -0.3954 & -0.3269\\
Top + Neg + Freq  & -0.2056 & -0.3608 & -0.3003\\
\hline
\end{tabular}
\label{tbl:correlations}
\end{table}

\subsection{Subversive Toxicity Detection}
Our second experiment consists in studying the benefits of taking sentiments into account when trying to determine whether a comment is toxic or not. The toxicity detector we implemented in this experiment is a deep neural network inspired by the most successful systems in the Kaggle toxicity competition we used as a dataset. It uses a bi-GRU layer with kernel size of 40. The final state is sent into a single linear classifier. To avoid overfitting, two 50\% dropout layers are added, one before and one after the bi-GRU layer. 

The network takes as input a sentence split into words and into individual characters. The words are represented by the 300d fastText pre-trained word embeddings\footnote{https://github.com/facebookresearch/fastText/blob/master /pretrained-vectors.md}, and characters are represented by a one-hot character encoding but restricted to the set of 60 most common characters in the messages to avoid the inclusion of noise. Finally, we used our "top + frequency" sentiment classifier with the average of the best three lexicons (SentiWordNet, Afinn, and Bing Liu) to determine the sentiment of each message. We input that information into the neural network as three sentiment values, corresponding to each of the three lexicons used, for each of the frequent words retained for the message. Words that are not among the selected frequent words or that are not found in a lexicon receive a sentiment input value of 0. Likewise, experiments that do not make use of sentiment information have inputs of 0 for all words. These input values are then concatenated together into a vector of 363 values, corresponding to the 300 dimensions of fastText, the 60 one-hot character vector, and the 3 sentiment lexicons.

The output of our network is a binary "toxic or non-toxic" judgment for the message. In the Kaggle dataset, this corresponds to whether the "toxic" label is active or not. In the Reddit dataset, it is the set of messages evaluated at levels 5, 6 or 7 by \textit{Community Sift} in any of the topics mentioned earlier. And in the Wikipedia dataset, it is any message marked as toxic by 5 workers or more. We chose this binary approach to allow the network to learn to recognize toxicity, as opposed to types of toxic messages on Kaggle, keyword severity on Reddit, or a particular worker's opinions on Wikipedia. However, this simplification created a balance problem: while the Reddit dataset is composed of 12\% toxic messages and 88\% non-toxic messages, the Wikipedia dataset is composed of 18\% toxic messages and the Kaggle dataset of 10\% toxic messages. To create balanced datasets, we kept all toxic messages and undersampled randomly the set of non-toxic messages to equal the number of toxic messages.

Our experiment consists in comparing the toxicity detection accuracy of our network when excluding or including sentiment information and in the presence of subversion. Indeed, as mentioned in Sections \ref{sec:introduction} and \ref{sec:background}, it is trivial for a subversive user to mask toxic keywords to bypass toxicity filters. In order to simulate this behavior and taking ideas from \cite{subversion}, we created a substitution list that replaces popular toxic keywords with harmless versions. For example, the word "kill" is replaced by "kilt", and "bitch" by "beach". Our list contains 191 words, and its use adds noise to $82\%$ of the toxic Kaggle messages, $65\%$ of the Wikipedia messages, and $71\%$ of the Reddit messages. These substitutions are only done at testing time, and not taken into account in training, to simulate the fact that users can create never-before-seen modifications.

\begin{table}
\caption{Accuracy of toxicity detection with and without sentiment}
\begin{tabular}{l|c|c}
\hline
Dataset & Without sentiment & With sentiment\\
 \hhline{=|=|=}
Kaggle &  93.2\%  & 93.7\%\\ 
Subversive Kaggle &  77.2\%  & 80.1\% \\
Wikipedia & 88.1\% & 88.5\%\\
Subversive Wikipedia & 81.4\% & 82.0\%\\
Reddit& 94.2\% & 94.3\% \\
Subversive Reddit& 83.0\% & 83.9\% \\
\hline
\end{tabular}
\label{tbl:toxicity}
\end{table}

We trained and tested our neural network with and without sentiment information, with and without subversion, and with each corpus three times to mitigate the randomness in training. In every experiment, we used a random 70\% of messages in the corpus as training data, another 20\% as validation data, and the final 10\% as testing data. The average results of the three tests are given in Table \ref{tbl:toxicity}. It can be seen that sentiment information helps improve toxicity detection in all cases. The improvement is smaller when the text is clean. However, the introduction of subversion leads to an important drop in the accuracy of toxicity detection in the network that uses the text alone, and the inclusion of sentiment information gives an important improvement in that case. Comparing the different corpora, it can be seen that the improvement is smallest in the Reddit dataset experiment, which is expected since it is also the dataset in which toxicity and sentiment had the weakest correlation in Table \ref{tbl:correlations}.

We can note that the system performs very well in all cases, even with subversion and without sentiment information. This may be due to the fact that the messages in all datasets are user-generated and therefore noisy already. In addition, the character encoding of the neural network is robust to misspellings, as opposed to a keyword lookup system.

\section{Conclusion}
\label{sec:conclusion}
In this paper, we explored the relationship between sentiment and toxicity in social network messages. We began by implementing a sentiment detection tool using different lexicons and different features such as word frequencies and negations. This tool allowed us to demonstrate that there exists a clear correlation between sentiment and toxicity. Next, we added sentiment information to a toxicity detection neural network, and demonstrated that it does improve detection accuracy. Finally, we simulated a subversive user who attempts to circumvent the toxicity filter by masking toxic keywords in their messages, and found that using sentiment information improved toxicity detection by as much as 3\%. This confirms our fundamental intuition, that while it is possible for a user to mask toxic words with simple substitutions, it is a lot harder for a user to conceal the sentiment of a message. 

Our work so far has focused on single-line messages and general toxicity detection. There are however several different types of toxicity, some of which correlate to different sentiments. For instance, while cyber-bullying and hate speech have negative sentiments, other forms of toxicity such as fraud or sexual grooming will use more positive sentiments in order to lure victims. We expect that differentiating between these types of toxicity will strengthen the correlation to message sentiment and further improve our results. Likewise, handling entire conversations instead of individual messages will allow us to include contextual information to better model the sentiment of the message, and to detect sudden changes in the sentiment of the conversation that may correspond to a disruptive toxic comment. 

\section*{Acknowledgements}
\label{sec:ack}
This research was made possible by the financial, material, and technical support of Two Hat Security Research Corp, and the financial support of the Canadian research organization MITACS.

\bibliographystyle{plain}
\bibliography{main}

\end{document}